\documentclass[runningheads]{llncs}

\usepackage{eccv}

\usepackage{eccvabbrv}

\usepackage{wrapfig}
\usepackage{graphicx}
\usepackage{booktabs}
\usepackage{multirow}
\usepackage{amsmath,amssymb}
\usepackage{array}
\usepackage{float}
\usepackage{caption}
\usepackage{amssymb}
\usepackage[table]{xcolor}  
\usepackage{arydshln}       
\usepackage{tcolorbox}
\usepackage{microtype}
\usepackage{tikz}
\usepackage{pgfplots}
\definecolor{gray1}{RGB}{135,175,210}
\definecolor{gray2}{RGB}{190,215,235}
\definecolor{gray3}{RGB}{225,235,245}
\usepackage{lmodern}   
\usepackage[T1]{fontenc}
\usepackage[accsupp]{axessibility}  
\usepackage{algorithm}
\usepackage{algpseudocode}
\usepackage[percent]{overpic}
\usepackage{csquotes}

\usepackage[pagebackref,breaklinks,hidelinks]{hyperref}
\hypersetup{
  colorlinks=true,
  linkcolor=red,      
  citecolor=blue,    
  urlcolor=blue,      
  filecolor=magenta,  
  anchorcolor=black,  
}

\usepackage{orcidlink}
\usepackage[disable]{todonotes}

\makeatletter
\renewcommand\paragraph{\@startsection{paragraph}{4}{\z@}%
  {.5ex \@plus .2ex \@minus .2ex}%
  {-1em}%
  {\normalfont\normalsize\bfseries}}
\makeatother

\newif\ifsupp
\suppfalse   

\begin{document}

\title{Dense Dynamic Scene Reconstruction and Camera Pose Estimation from Multi-View Videos}

\titlerunning{Abbreviated paper title}

\authorrunning{Shuo Sun et al.}
\titlerunning{Dense Dynamic Reconstruction from Multiple Videos}
\author{Shuo Sun\inst{1}\orcidlink{0000-0003-0790-6510}
\and
Unal Artan\inst{1}\orcidlink{0000-0002-6730-9278
}
\and
Malcolm Mielle\inst{2}\orcidlink{0000-0002-3079-0512}
\and
Achim J. Lilienthal\inst{1,3}\orcidlink{0000-0003-0217-9326}
\and
Martin Magnusson\inst{1}\orcidlink{0000-0002-8049-2142}
}


\institute{
Örebro University, Sweden \\
\and
Schindler EPFL Lab, Lausanne, Switzerland \\
\and
Technical University of Munich, Germany \\
}

\maketitle

\ifsupp
\else
\begin{center}
    \centering
    \captionsetup{type=figure}
\includegraphics[width=\textwidth]{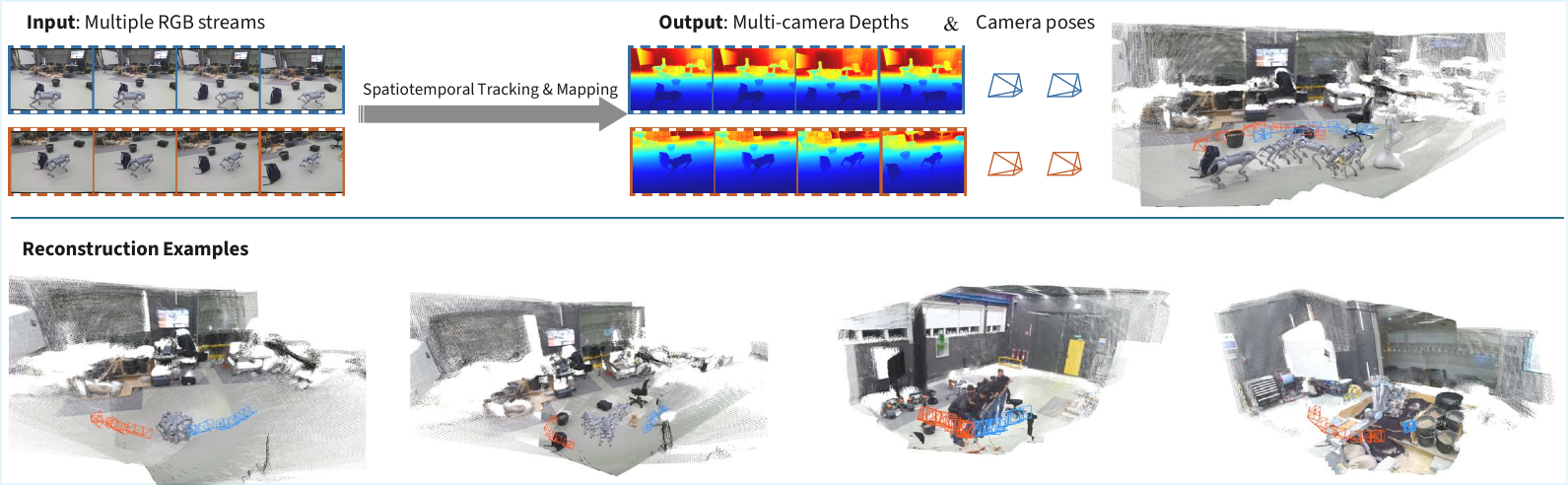}
    \captionof{figure}{Given video sequences
    captured from multiple free cameras, our method can recover dense dynamic scenes
    consistently and estimate camera poses accurately.
    We illustrate our complete pipeline (top row) together with reconstruction results on four additional sequences (bottom row), where the red and blue frames indicate cameras.
    \label{fig:grphic_abstract}}
    \vspace{-4pt}
\end{center}%

\begin{abstract}
    We address the challenging problem of dense dynamic scene reconstruction and camera pose estimation from multiple
    freely moving cameras---a setting that arises naturally when multiple observers capture a shared event.
    Prior approaches either handle only single-camera input or require rigidly mounted, pre-calibrated camera rigs, limiting their practical applicability.
    We propose a two-stage optimization framework that decouples the task into robust camera tracking and dense depth refinement.
    In the first stage, we extend single-camera visual SLAM to the multi-camera setting by constructing a spatiotemporal connection graph that exploits both intra-camera temporal continuity and inter-camera spatial overlap, enabling consistent scale and robust tracking.
    To ensure robustness under limited overlap, we introduce a wide-baseline initialization strategy using feed-forward reconstruction models.
    In the second stage, we refine depth and camera poses by optimizing dense inter- and intra-camera consistency using wide-baseline optical flow.
    Additionally, we introduce MultiCamRobolab, a new real-world dataset with ground-truth poses from a motion capture system.
    Finally, we demonstrate that our method significantly outperforms state-of-the-art feed-forward models on both synthetic and real-world benchmarks, while requiring less memory.
    Our code is released at \href{https://github.com/ljjTYJR/Multiple-view-Dynamic-Reconstruction}{\texttt{https://github.com/ljjTYJR/Multiple-view-Dynamic-Reconstruction}}.
\end{abstract}
\section{Introduction}
\label{sec:intro}
Consider a scenario in which multiple observers simultaneously capture a dynamic scene from diverse viewpoints.
Accurately reconstructing dynamic scenes from these observations is challenging due to temporal dynamics and limited view overlap.
In this work, we study the problem of dynamic scene reconstruction from multiple free cameras: 
\textit{Given multiple video inputs, 
how can we recover dense, dynamic scenes at any time step and estimate camera poses at the same time?}

Multi-camera capture is increasingly common in robotics~\cite{heng2019project}, sports broadcasting~\cite{yus2015multicamba}, and consumer devices where multiple phones or action cameras record a shared event.
Applications such as augmented reality, dynamic Gaussian splatting~\cite{azzarelli2025splatography,wang2025monofusion}, and multi-view video analysis~\cite{li2025multi} all benefit from accurate, consistent 3D reconstruction across cameras.
Yet existing dynamic scene reconstruction methods are limited to single-camera setups~\cite {zhang2024monst3r,lu2025align3r,li2025megasam,wang2025shape,huang2025vipe}.
For instance, MegaSAM~\cite{li2025megasam}---a robust monocular visual SLAM---exploits only temporal connections within a single camera, leaving cross-camera consistency unexploited.
While multi-camera SLAM methods exist~\cite{lajoie2020door, tian2022kimera, hu2023cp, yugay2025magic,heng2019project}, they either assume rigid camera rigs with known calibration or focus on static global map construction, failing to address dynamic scenes with freely moving cameras.

The key technical challenges are threefold:
(1) \textit{Scale ambiguity}: monocular depth is inherently scale-ambiguous, and without shared observations, each camera's reconstruction may drift to a different scale;
(2) \textit{Limited overlap}: unlike a rigid rig, free-moving cameras may have minimal or intermittent view overlap, complicating inter-camera constraints;
(3) \textit{Dynamic content}: moving objects violate the static-world assumption underlying classical multi-view geometry, requiring robust correspondence estimation.

To address these challenges, we introduce a two-stage optimization framework.
In the first stage, we extend single-camera SLAM to multiple cameras by constructing a \textit{spatio-temporal connection graph} that links frames across cameras based on view overlap, enabling joint bundle adjustment with consistent scale.
To overcome the challenges of multi-camera initialization under limited overlap, we introduce a wide-baseline initialization strategy using a feed-forward reconstruction model. This approach provides a unified scale anchor and provides a reliable starting point for subsequent pose optimization.
In the second stage, with coarse camera poses obtained, we refine per-frame depth and camera poses by optimizing dense inter- and intra-camera consistency using dense wide-baseline optical flow correspondences.

Our contributions are as follows: 1) We propose a multi-camera tracking framework that can achieve consistent dynamic scene reconstruction from multiple free-moving cameras. To the best of our knowledge, our work is the first one designed for this task.
2) We collect and make available a new dataset, MultiCamRobolab, enabling quantitative evaluation of methods for multi-camera dynamic scene reconstruction and camera pose tracking.
3) We demonstrate that our method achieves better tracking and reconstruction results compared to state-of-the-art methods while consuming less memory.

\section{Related work}
\label{sec:related_work}
Our method is related to visual SLAM and structure from motion(SfM), two well-established fields for 3D scene reconstruction and camera pose estimation.
However, to the best of our knowledge, no existing method explicitly targets the research question addressed in this paper, namely, multi-camera dynamic scene reconstruction.
Below, we highlight key advancements and limitations in the state of the art that motivate our contributions.

\paragraph{Visual SLAM and SfM}
Classical visual SLAM methods~\cite{mur2015orb,engel2014lsd, engel2017direct,klein2007parallel,homeyer2025droid} are primarily designed for single-view monocular camera configurations.
Many works extend this setting to multi-camera configurations.
One line of research employs calibrated camera rigs~\cite{heng2019project, schmied2023r3d3,kuo2020redesigning,liu2018towards}, in which multiple cameras are rigidly mounted with known extrinsic parameters. Each camera provides complementary views of the environment, enabling wider field-of-view coverage and improved robustness in challenging scenarios.
Another direction leverages multiple cameras in the context of collaborative (multi-agent or multi-session) SLAM~\cite{deng2025mne,9686955,yugay2025magic,duhautbout2019distributed, 10655847}, where each camera (or agent) maintains an independent local mapping process. The individually built static sub-maps are later fused together.
Our method differs from the above-mentioned approaches in two primary respects. First, unlike calibrated camera-rig systems, we allow cameras to move freely without requiring known inter-camera extrinsics. Second, in contrast to multi-agent SLAM frameworks, which typically assume static environments, our objective is to reconstruct dense and dynamic scenes.

Structure from motion (SfM) processes unordered image inputs, reconstructing the scene and estimating camera poses at the same time.
Though accurate, classical SfM methods~\cite{pan2024global, schonberger2016structure} only achieve sparse representation and are brittle to dynamic objects in the view.
On the other hand, recent feed-forward 3D reconstruction models~\cite{wang2025vggt,wang2024dust3r,yang2025fast3r}, trained on large-scale datasets, exhibit substantially improved robustness under sparse input conditions compared with classical SfM pipelines. 
Nevertheless, such models are typically memory-intensive and encounter difficulties when applied to long video sequences.
Though some methods~\cite{deng2025vggt} utilize such models in a chunk-wise way to handle long videos, their results are not as good as global predictions~\cite{ding2025laser}.

\paragraph{Dense dynamic scene reconstruction}
If given camera poses, the simplest way is to use monocular depth prediction models~\cite{piccinelli2024unidepth,bochkovskii2024depth,huang2025vipe} to predict per-frame depth.
However, the monocular depth model alone can often generate flickering and inconsistent results along a video~\cite{kopf2021robust}.
Previous works solve the problem either via consistent video depth optimization~\cite{kopf2021robust,luo2020consistent} or directly train video depth prediction~\cite{chen2025video, kuang2025buffer, hu2025depthcrafter}.
Recently, many works reconstructing dynamic scenes rely on \enquote{3D reconstruction foundation} models~\cite{feng2025st4rtrack,zhang2024monst3r,lu2025align3r,wang2025continuous,leroy2024grounding,wang2025pi} to predict 3D scenes and camera poses at the same time.
For example, \textit{Monst3R}~\cite{zhang2024monst3r} fine-tunes the Dust3R model on dynamic videos and optimizes camera poses for long-sequence videos.
\textit{CUT3R}~\cite{wang2025continuous} uses a stateful recurrent model to process video inputs, achieving significant memory consumption reduction compared to global attention computation.
However, existing feed-forward reconstruction models are either memory-intensive or produce low-quality reconstructions. In contrast, our method decouples camera pose estimation from depth optimization, enabling better reconstruction quality and scalability to long video sequences.
The most relevant work to ours is~\cite{mustafa2016temporally}, which reconstructs a mesh of dynamic objects in the view; however, it requires a fixed camera setup and prior extrinsic calibration. In contrast, our work can reconstruct scenes from freely moving cameras.

\section{Methods}
\label{sec:methods}
\subsection{Overview}
\paragraph{Problem Definition.}
As shown in \cref{fig:grphic_abstract}, given a set of time-synchronized monocular video sequences
\begin{equation*}
\mathcal{I} = \{ (\mathbf{I}_i^t, \mathbf{K}_i) \mid i = 1,\dots,N; \, t = 1,\dots,T \},
\end{equation*}
where $\mathbf{I}_i^t \in \mathbb{R}^{H \times W \times 3}$ denotes the image captured by the $i$th camera at timestamp $t$,
and $\mathbf{K}_i$ is the intrinsic parameters for the $i$th camera, the objective is to estimate the per-frame camera states
\begin{equation*}
\mathcal{X} = \{ (\mathbf{T}_i^t, \mathbf{D}_i^t) \mid i = 1,\dots,N; \, t = 1,\dots,T \},
\end{equation*}
where $\mathbf{T}_i^t \in SE(3)$ represents the camera pose
and $\mathbf{D}_i^t \in \mathbb{R}^{H\times W}$ denotes the corresponding depth map.

In the following section, we first introduce how we track multiple cameras accurately and robustly.
Then we demonstrate how to refine the scene consistently given the estimated camera poses.
\Cref{fig:overview} shows an overview of our approach.
\newcommand{\refsize}{\fontsize{3.2pt}{1.0pt}\selectfont}

\begin{figure}[t]
    \begin{tikzpicture}
      \node[anchor=south west, inner sep=0] (img) at (0,0)
        {\includegraphics[width=\textwidth]{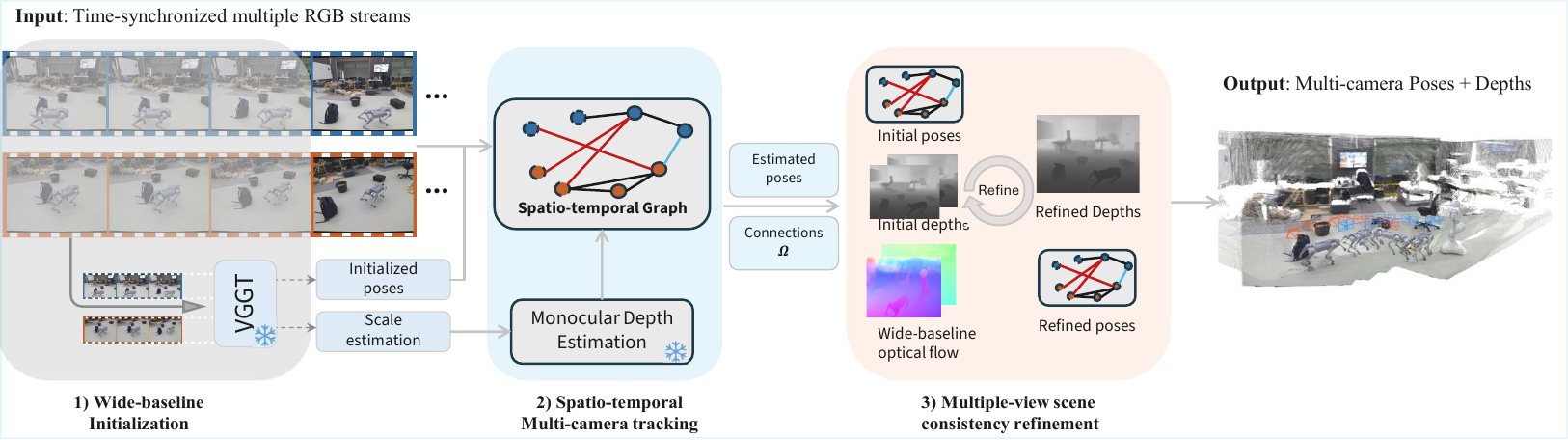}};
    
    
    
    
    
    \end{tikzpicture}
    \caption[Method Overview.]{
    \textbf{Method Overview.} Given multiple video inputs: 
    Our method first uses a feed-forward model for initialization to achieve a global scale anchor and initialized poses.
    Then, we build a spatio-temporal connection graph during tracking to estimate camera poses and maintain a consistent scale.
    At last, we leverage the dense optical flow, estimated poses, and achieved connection graph to refine per-pixel depth to get a consistent scene and refined camera poses.
    }
    
    \label{fig:overview}
    \vspace{-4pt}
\end{figure}

\subsection{Spatio-temporal multi-camera tracking}
\label{subsec:multi-cam tracking}
\paragraph{Preliminary.} We utilize the learned correspondence estimation model from MegaSAM~\cite{li2025megasam}. Given two frames $\mathbf{I}_i$ and $\mathbf{I}_j$,
MegaSAM learns to predict the dense optical flow $\mathbf{f}_{i \to j} \in \mathbb{R}^{H'\times W'\times 2}$ and weights $\mathbf{w}_{ij} \in \mathbb{R}^{H'\times W'}$ with lower resolution $(H' \times W')$ in an iterative way.
Given a grid of pixel coordinates $\mathbf{u}_{i} \in \mathbb{R}^{H'\times W'\times 2} $ in frame $\mathbf{I}_{i}$, we can predict the flow-warped correspondence of $\mathbf{u}_{i}$ on the image $\mathbf{I}_{j}$ by $\mathbf{u}_{ij}^{\text{flow}} = \mathbf{u}_{i} + \mathbf{f}_{i \to j}$.
With the current estimated state, we can also project $\mathbf{u}_i$ directly on the image $\mathbf{I}_j$ and achieve the reprojection $\mathbf{u}_{ij}^{\text{reproj}}$:
\begin{equation}
    \mathbf{u}_{ij}^{\text{reproj}} = \mathbf{K}_j(\mathbf{T}_{ij} \circ \mathbf{K}_i^{-1}(\mathbf{u}_i, \mathbf{d}_i)),
    \label{eq:projection}
\end{equation}
where $\mathbf{u}_{ij}^{\text{reproj}} \in \mathbb{R}^{H'\times W'}$ is the reprojected correspondence of $\mathbf{u}_{i}$ on frame $\mathbf{I}_j$; $\mathbf{T}_{ij} = \mathbf{T}_j^{-1} \circ \mathbf{T}_i$; $\mathbf{d}_i$ is the disparity map of frame $\mathbf{I}_i$.
Given a group of connected frames $\mathbf{\Omega}$, parameters $\mathbf{T}$ and $\mathbf{d}$ are optimized by minimizing the weighted re-projection error:
\begin{equation}
    \mathcal{L}(\mathbf{T},\mathbf{d}) = \sum_{(i,j)\in \mathbf{\Omega}}||\mathbf{u}_{ij}^{\text{reproj}} - \mathbf{u}_{ij}^{\text{flow}}||_{\Sigma_{ij}}^{2}
    \label{eq:ba}
\end{equation}
where $\Sigma_{ij}={\texttt{diag}}(\mathbf{w}_{ij})$ are weights, where possible dynamic objects will be assigned low weights to reduce their effects.

\paragraph{Spatio-temporal connection graph $\Omega$.}\label{para:spatio-temporal}
To extend the monocular tracking framework to multiple cameras, one of our key contributions is to introduce a \textit{spatio-temporal connection graph}.
Our frame connection graph consists of three parts.
\textbf{Temporal connection} $\mathbf{\Omega}^{\rm{temp}}$:
For each single camera, since its frames are sequential in time, adjacent frames tend to have enough overlap for reliable correspondence estimation.
In this case, we follow common practice in one single-camera setting, maintaining a temporal window to hold the latest keyframes for each camera,
which means $\mathbf{\Omega}^{\rm{temp}} = \bigcup_{i} \mathbf{\Omega}_{i}^{\rm{temp}}$ for each $i$-th camera.
\textbf{Spatial connection} $\mathbf{\Omega}^{\rm{spat}}$: In addition to the temporal intra-connection for each single camera, the spatial inter-connection is also necessary to maintain the scale consistency and improve accuracy.
Given the $N$ frames from all cameras at timestamp $t$, we decide the inter-camera connection by evaluating the overlap across different camera frames: given two RGB frames $\mathbf{I}_i^{t}$ and $\mathbf{I}_j^{t}$, we project the grid of coordinates $\mathbf{u}_i$ to the frame $\mathbf{I}_j^{t}$ by \cref{eq:projection} to achieve the corresponding pixels $\mathbf{u}_{ij}^{t}$ on the frame $\mathbf{I}_j$'s image plane.
If most of the projected pixels (more than 75\%) are within the boundary of the image, then we make a spatial connection.
This is helpful because even though there are dynamic objects in the scene, at the same timestamp, from different views, the scene is still consistent in this instant.
\textbf{Spatio-temporal connection} $\mathbf{\Omega}^{\text{st}}$: In addition to the intra-camera temporal connection and inter-camera spatial connection at timestamp $t$, we also exploit the historical information across different cameras.
At timestamp $t_0$, keyframe $\mathbf{I}_i^{t_0}$ will try to evaluate $\mathbf{u}_{ij}^{t_0}$ with all other cameras' inactive keyframes $\{\mathbf{I}_j^{t}|j=[1,N], j\neq i, t=[1,T'], T' < t_0\}$.
If there is enough overlap, we also make a connection between two frames.
To prevent memory explosion and make connections effectively while running, we implement a \textit{connection balance} strategy: we set a maximum number of edges in the tracking active window and allocate inter-camera connections evenly if they exist.
If newly detected connections plus the existing connections are more than the maximum number, we remove the oldest ones from the tracking active window.
A graphic demonstration of the spatio-temporal graph can be found in \cref{fig:spatiotemporal-graph}.

\begin{wrapfigure}{r}{0.5\textwidth}
    \centering
    \includegraphics[width=\linewidth]{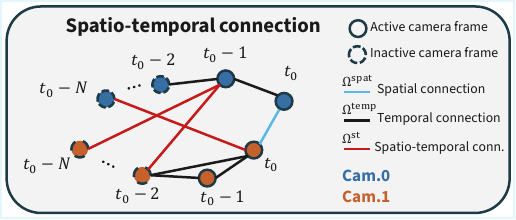}
    \caption{\textbf{Demonstration spatio-temporal graph.} First, each single camera will estimate temporal connections with its own frames. Second, at the timestamp $t_0$, {\color{RedOrange}\textbf{Cam.1}} will try to make a spatial connection with {\color{MidnightBlue}\textbf{Cam.0}} if there is enough overlap.
    Additionally, the current active keyframe will try to make spatio-temporal connections with those inactive frames from other cameras if there is enough overlap.
    Ablation studies~(\cref{subsec:ablation_study}) show spatio-temporal connections improve tracking accuracy.
    }
    \label{fig:spatiotemporal-graph}
    \vspace{-4pt}
\end{wrapfigure}
In all, during multi-camera tracking, we create the spatio-temporal connection graph
\begin{equation*}
\mathbf{\Omega} = \mathbf{\Omega}^{\text{temp}} \cup \mathbf{\Omega}^{\text{spat}} \cup \mathbf{\Omega}^{\text{st}}.
\end{equation*}
We prove the effectiveness of the spatio-temporal connection graph in \cref{subsec:ablation_study}.

\paragraph{Wide-baseline Initialization.}\label{para:init}
In one single-camera setting, tracking initialization can typically rely on selecting frames with sufficient inter-frame overlap, which is naturally satisfied due to the temporal continuity of the image stream. In contrast, in multi-camera configurations, it is common that images captured from different viewpoints have limited or even no overlap at all, which makes conventional overlap-based initialization unreliable.

To address this issue, we adopt the robust feed-forward scene reconstruction model VGGT~\cite{wang2025vggt} to initialize our system. Specifically, during initialization, we select the first $N_{\text{init}}$ frames from each camera stream and input them into VGGT to obtain initial camera pose estimates along with per-frame depth predictions $\mathbf{D}_{\text{VGGT}}$. These predictions provide a coarse but globally consistent geometric initialization across multiple cameras.

To further enhance tracking robustness and to obtain dense depth priors, similar to MegaSAM, we additionally incorporate monocular depth estimates during tracking. Concretely, we employ a monocular depth estimation model, \textit{UniDepth}~\cite{piccinelli2024unidepth}, to predict per-frame monocular depth maps $\mathbf{D}_{\text{mono}}$. Since monocular depth is only defined up to an affine transformation, we align the predicted depths to the initialization scale by estimating a global scale $s \in \mathbb{R}$ and offset $o \in \mathbb{R}$ through optimizing
\begin{equation}
s, o = \min \sum_{k}||s \mathbf{D}^{k}_{\text{mono}} + o - \mathbf{D}_{\text{VGGT}}^{k}||^{2}, \label{scale:equation}
\end{equation}
where $k$ indexes the selected initialization frames and $\mathbf{D}^{k}_{\text{VGGT}}$ denotes the corresponding depth predictions from VGGT. The estimated scale and offset are then applied consistently to all monocular depth predictions during tracking.

After obtaining the initial camera poses and depth predictions from the feed-forward model VGGT, 
we run 
bundle adjustment to further refine the disparity maps and poses.
Through the wide-baseline initialization, the system is able to establish a robust and metrically consistent initialization across multiple cameras from the outset, which significantly stabilizes subsequent tracking and optimization.

\paragraph{Tracking system.}\label{tracking:para}
After initialization, we incrementally construct the spatio-temporal connection graph $\mathbf{\Omega}$ and jointly optimize camera poses and per-frame depths. As discussed in the \hyperref[para:init]{Wide-baseline Initialization} paragraph, to enhance robustness and enforce scene-scale consistency, we incorporate a prior depth regularization term into the bundle adjustment during tracking.

As described in the \hyperref[para:spatio-temporal]{Spatio-temporal connection graph}, upon the arrival of new frames $\{\mathbf{I}_i\}_{i}^{N}$, each frame is associated with previously processed frames to estimate new spatio-temporal connections, which are then added to $\mathbf{\Omega}$. The camera pose of each new frame is initialized under a constant-velocity motion assumption:
$\mathbf{T}_i^{t} \leftarrow (\mathbf{T}_i^{t-1} \cdot (\mathbf{T}_{i}^{t-2})^{-1})$.
To further improve tracking robustness, we integrate aligned monocular depth predictions as priors in the bundle adjustment. The resulting optimization problem is formulated as
\begin{equation}
    \mathcal{L}(\mathbf{T},\mathbf{d}) = \sum_{(i,j)\in \mathbf{\Omega}}||\mathbf{u}_{ij}^{\text{reproj}} - \mathbf{u}_{ij}^{\text{flow}}||_{\Sigma_{ij}}^{2} + \lambda || \mathbf{d}_i - \mathbf{D}_{i}^{\text{s}}||^{2}
    \label{eq:final_ba}
\end{equation}
The reference depth $\mathbf{D}_i^{\mathrm{s}}$ is obtained by aligning the monocular depth prediction to the initialization scale via $\mathbf{D}_{i}^{\text{s}}= s \mathbf{D}^{i}_{\text{mono}} + o$, where the global scale $s$ and offset $o$ are estimated as described in \cref{scale:equation}.
For online optimization, we maintain an active sliding window consisting of the most recent frames. Within this window, the poses of the earliest frames are fixed to remove the gauge freedom.

\subsection{Multiple-view scene consistency refinement}
\label{subsec:refinement}
The tracking stage (\cref{subsec:multi-cam tracking}) produces initial camera pose and depth estimates:
$$
\mathcal{X}_{\text{init}} = \{ (\mathbf{T}_i^t, \mathbf{D}_i^t) \mid i = 1,\dots,N; \, t = 1,\dots,T \},
$$
where $\mathbf{D}_i^t = s \mathbf{D}^{i,t}_{\text{mono}} + o$ represents the affine-aligned monocular depth from \cref{scale:equation}.

While this global affine alignment ensures metric scale consistency at initialization, it is insufficient for achieving dense multi-view geometric consistency throughout the full sequence.
Specifically, the simple affine transformation cannot account for:
(1) per-frame scale drift in monocular depth predictions across the video sequence, and
(2) per-pixel depth inaccuracies from the monocular depth model prediction.
Therefore, we perform a multi-view depth refinement stage that optimizes per-frame scale and per-pixel depth corrections across all cameras simultaneously.

Depth refinement from monocular video has been extensively studied in prior work~\cite{kopf2021robust,luo2020consistent,li2025megasam}.
In this section, we extend these techniques to the multi-camera setting, exploiting both temporal consistency within each camera and spatial consistency across different cameras.

\paragraph{Dense correspondence estimation.}
To enable robust multi-view optimization, we compute dense optical flow correspondences that go beyond the low-resolution flow used during tracking.
We construct an augmented connection graph $\mathbf{\Omega}_{\text{refine}}$ that includes: (1) the spatio-temporal graph $\mathbf{\Omega}$ from tracking (\cref{subsec:multi-cam tracking}), and (2) additional dense temporal connections within each camera sequence.
Specifically, motivated by prior monocular video depth optimization~\cite{kopf2021robust,luo2020consistent,li2025megasam}, for each frame $i$ in a camera's sequence, we connect it to frames at offsets $\{+2, +4, +8\}$ when they exist, enabling longer-range temporal consistency constraints.

For all frame pairs in $\mathbf{\Omega}_{\text{refine}}$, we estimate dense optical flow using the wide-baseline flow estimation model UFM~\cite{zhang2025ufm}, which provides higher quality correspondences than the tracking-stage flow, particularly for large inter-frame motion and inter-camera baselines.





\paragraph{Optimization objectives.}
Let $(i, j)$ denote a connected frame pair in $\mathbf{\Omega}_{\text{refine}}$.
For brevity, we omit the timestamp superscript $t$ in the following derivations.

Our optimization minimizes the weighted reprojection error:
\begin{equation}
\mathcal{L}_{\text{reproj}} = w_f \mathcal{L}_{\text{flow}} + w_d \mathcal{L}_{\text{disp}} \label{eq:depth-opt}
\end{equation}

The flow reprojection term $\mathcal{L}_{\text{flow}}$ penalizes misalignment between optical flow correspondences and geometric reprojections.
Specifically,
{\small
\begin{equation}
\mathcal{L}_{\text{flow}} = \frac{\sum_{(i,j)} \left( \|\mathbf{u}_{ij}^{\text{reproj}} - \mathbf{u}_{ij}^{\text{flow}}\|_1 \cdot \mathbf{c}_i + \lambda \log\frac{1}{\mathbf{c}_i} \right) \cdot m_{ij}}{ \sum_{(i,j)} m_{ij}},
\end{equation}
}%
where $\mathbf{u}_{ij}^{\text{flow}} = \mathbf{u}_i + \mathbf{f}_{i \to j}$ is the optical flow correspondence, $\mathbf{c}_i \in (0,1]$ is an optimizable per-flow confidence map, and $m_{ij}$ is the valid flow mask.
The $\log(1/\mathbf{c}_i)$ term prevents the confidence from collapsing to zero.

The geometric reprojection $\mathbf{u}_{ij}^{\text{reproj}}$ is computed as:
\begin{equation}
\mathbf{u}_{ij}^{\text{reproj}} = \mathbf{K}_j(\mathbf{T}_{ij} \circ \mathbf{K}_i^{-1}(\mathbf{u}_i, s_i \mathbf{D}_i + \beta_i)),
\end{equation}
where $s_i \in \mathbb{R}$ and $\beta_i \in \mathbb{R}$ are per-frame scale and shift parameters to be optimized.
Note that these per-frame parameters are independent of the global initialization scale $(s, o)$ from \cref{scale:equation}.
The disparity consistency term $\mathcal{L}_{\text{disp}}$ penalizes inconsistencies between forward-projected and estimated depths, helping to reduce depth flickering~\cite{luo2020consistent}:
{\small
\begin{align}
    \mathcal{L}_{\text{disp}} = \frac{\sum_{(i,j)}  (\left| \frac{\max(\mathbf{D}_j^{\text{flow}}, \mathbf{D}_j)}{\min(\mathbf{D}_j^{\text{flow}}, \mathbf{D}_j)} - 1 \right| \cdot \mathbf{c}_i +  \lambda \log\frac{1}{\mathbf{c}_i}) \cdot m_{ij}}{\sum_{(i,j)} m_{ij}}
\end{align}
}%
where $\mathbf{D}_j^{\text{flow}}$ is the depth warped from frame $i$ to frame $j$ via optical flow, while $\mathbf{D}_j$ is the estimated depth at frame $j$.

To improve the optimization stability, we refine the depth and camera poses in two phases.
\paragraph{Phase 1: Per-frame scale alignment.}
\label{para:phase1}
In the first phase, we fix all camera poses $\{\mathbf{T}_i^t\}$ to their initial estimates from tracking and optimize the per-frame affine parameters $\{s_i, \beta_i\}$ along with the flow confidence maps $\{\mathbf{c}_i\}$ by minimizing $\mathcal{L}_{\text{reproj}}$ (\cref{eq:depth-opt}).
This establishes a consistent scale across all frames while down-weighting unreliable flow correspondences through learned confidence.

\paragraph{Phase 2: Iterative pose and depth refinement.}
\label{para:phase2}
Phase 1 establishes frame-level scale consistency but does not correct per-pixel depth errors in the monocular predictions.
In Phase 2, we fix per-frame affine parameters $(s_i, \beta_i)$, and directly optimize the per-pixel depth values $\mathbf{D}_i$ and refine camera poses by minimizing the same reprojection loss $\mathcal{L}_{\text{reproj}}$.
Since jointly optimizing camera poses and depths suffers from instability~\cite {li2025megasam}, we propose to optimize poses and depths in an alternating iterative manner.
For camera pose optimization, we augment the reprojection loss with pose regularization terms to maintain temporal smoothness:

Let the camera pose perturbation be
$\boldsymbol{\delta}_i = [\omega_x, \omega_y, \omega_z, t_x, t_y, t_z]^{\top}  \in \mathbb{R}^6,$
where $[\omega_x, \omega_y, \omega_z]^\top$ is the rotation in axis-angle form and $[t_x, t_y, t_z]^\top$ is the translation.

The pose update $\Delta_i = \exp(\boldsymbol{\delta}_i) \in SE(3)$ yields the updated pose:
\begin{equation}
\mathbf{T}_i^{\text{new}} = \mathbf{T}_i^{\text{orig}} \cdot \exp\left(\begin{bmatrix} [\boldsymbol{\omega}_i]_\times & \mathbf{t}_i \\ 0 & 0 \end{bmatrix}\right)
\end{equation}
The pose regularization loss consists of two components:
$\mathcal{L}_{\text{pose}} = \mathcal{L}_{\text{prior}} + \mathcal{L}_{\text{smooth}}$,
where $\mathcal{L}_{\text{prior}} = w_{\text{prior}} \sum_i||\boldsymbol{\delta}_{i}||^{2}$ penalizes large deviations from the initial pose estimates.
The smoothness term $\mathcal{L}_{\text{smooth}} = \mathcal{L}_{\text{smooth}}^{\text{rot}} + \mathcal{L}_{\text{smooth}}^{\text{trans}}$ enforces temporal consistency along each camera trajectory, where $\mathbf{T}_i^t = \begin{bmatrix} R_t^{(i)} & \mathbf{t}_t^{i} \\ 0 & 1 \end{bmatrix}$ with $R_t^{(i)} \in SO(3)$ and $\mathbf{t}_t^{i} \in \mathbb{R}^3$ being the rotation and translation components.
The smoothness terms are defined as:
$$
\mathcal{L}_{\text{smooth}}^{\text{rot}}(\boldsymbol{\delta}) = w_{\text{tem-rot}} \sum_{i=1}^{N_{\text{cam}}} \sum_{t=1}^{N_i-1} \left\| R_t^{(i)T} R_{t+1}^{(i)} - I \right\|_F^2,
$$
and
$$
\mathcal{L}_{\text{smooth}}^{\text{trans}} = w_{\text{tem-trans}} \sum_{i=1}^{N_{\text{cam}}} \sum_{t=1}^{N_i-1}  \left\| \mathbf{t}_{t}^{i} - \mathbf{t}_{t+1}^{i}\right\|^{2}.
$$
The overall pose optimization loss is $\mathcal{L} = \mathcal{L}_{\text{reproj}} +\mathcal{L}_{\text{pose}}$.

\section{Real-World Multi-Camera Dataset}

To assess our performance in real-world settings, we introduce the \textit{MultiCamRobolab} \footnote{\url{https://github.com/ljjTYJR/multiple-view-dynamic-reconstruction}}.
This dataset comprises 24 RGB-D sequences acquired in a laboratory environment using two or three Microsoft Azure Kinect cameras under diverse motion patterns. Our method only requires RGB images. The depth information from the RGB-D Azure Kinects is only used to quantify performance.
Ground-truth camera poses are provided by a Qualisys motion-capture system.
Temporal synchronization between the RGB cameras and the Qualisys system is achieved via a time server running on the Qualisys PC.
Specifically, timestamps are recorded for image frames and motion-capture measurements, and pairs whose timestamp differences are smaller than the sampling interval are regarded as synchronized observations.
Each video clip is captured at 30 FPS and consists of 150–300 frames.
The 24 sequences are divided into 5 distinct scenarios.
The first 19 sequences are collected with 2 cameras: 1) $\text{Robodog}_\text{overlap}$ (4 sequences), where a robot dog is walking in the scene and two cameras have enough view overlap; 2) $\text{Robodog}_\text{non-overlap}$ (4 sequences), similar to the first case, but two cameras have little or no overlap;
3) RoboArm (4 sequences), where a 6-DOF manipulator performs digging motions, and 4) DynamicHuman (7 sequences), where a human is walking or working in the scene.
The last 5 sequences are collected with 3 cameras:
5) Three-camera (5 sequences), where a human operator is operating, or a robot dog is moving in the scene.


\section{Experiments setting}

\subsection{Baseline methods}
We select \textit{COLMAP}~\cite{schonberger2016structure} and \textit{GLOMAP}~\cite{pan2024global} as the baseline to represent the classical optimization-based structure-from-motion method.
Since COLMAP and GLOMAP only generate sparse point models, we limit our comparison to camera pose evaluation when using them.
In experiments, we set the known camera intrinsics for COLMAP and GLOMAP.
In addition to the default settings, we also test COLMAP/GLOMAP with two extra settings for fair comparison: \enquote{with dynamic masks}, which prunes dynamic objects from the views, and \enquote{with NetVLAD retrieval}, which uses NetVLAD for image retrieval to establish more image connections.

We also evaluate against recent state-of-the-art feed-forward models: \textit{Fast3R}~\cite{yang2025fast3r},
\textit{VGGT}~\cite{wang2025vggt}, and a memory-efficient version of VGGT \textit{FastVGGT}~\cite{shen2025fastvggt}. These methods demonstrate strong performance in scenarios where traditional methods often fail.
Finally, we include \textit{CUT3R}~\cite{wang2025continuous} as a baseline due to its specific design for video processing, while it also supports unstructured image inputs.
When running CUT3R, we sequentially feed frames from each camera video (i.e., camera-1 followed by camera-2) to fully leverage its temporal processing capabilities.

Fast3R and FastVGGT are evaluated on a single NVIDIA A100 GPU with 40 GiB of memory due to their higher memory requirements, whereas all other methods are tested on a single NVIDIA RTX 4090 GPU.
Since VGGT cannot process all frames on a single A100 GPU due to memory limitations, we subsample the input sequence with an interval of 8 during evaluation.

\subsection{Metrics}

\paragraph{Camera Pose Evaluation.}\label{para:camera_pose_eva}
First, camera poses are aligned using the initial ground truth pose to maintain a common reference coordinate system.
We then assess the cameras' trajectory accuracy with standard error metrics: Absolute Translation Error (ATE), Relative Translation Error (RTE), and Relative Rotation Error (RRE).
Note that we assess the multiple camera trajectories as a single trajectory.
The translation error is measured in meters, while the rotation error is measured in degrees.

\paragraph{Depth Quality Assessment.}\label{para:depth_quality_eva}
We evaluate depth quality using  Absolute Relative Depth (Abs.Rel) and  Delta accuracy~\cite{li2025megasam,zhang2024monst3r}($\delta<1.25$, i.e., the percentage of predicted depths within a 1.25-factor of true depth).
Per-frame depth evaluation is conducted by scaling the predicted depth and offset to the ground truth.

\paragraph{Scene Consistency.}\label{para:scene_cons_eva}
Since depth metrics alone do not capture scene-level consistency, we evaluate scene consistency using point-to-point distance as the relevant metric. First, we align the predicted trajectory with the ground truth to compute a global scale factor. The predicted depth maps are subsequently scaled using this factor and re-projected into 3D coordinates. Consistency is then measured by computing the Euclidean distance between corresponding 3D points in the predicted and ground-truth reconstructions.  The median Euclidean distance (denoted as $M_d$) is reported to minimize the impact of outliers.

\subsection{Datasets}
In addition to our self-collected MultiCamRobolab dataset,
we also evaluate our method against the MultiCamVideo-Dataset~\cite{bai2025recammaster}.
MultiCamVideo-Dataset is a multi-camera synchronized video dataset rendered in simulation using Unreal Engine, which includes synchronized multi-camera videos and their corresponding camera trajectories. Each video clip has 81 frames with 10 different camera perspectives.
In our evaluation, we randomly extract 50 clips and 3 random cameras for each clip in the MultiCamVideo-Dataset. All images are cropped to $512\times384$, and trajectories are normalized in $[-1,1]$.

\section{Results}
\subsection{Quantitative \& Qualitative Comparisons}
\begin{table}[t]
\centering
\caption{Quantitative comparisons of camera trajectories on the \textbf{MultiCamVideo}~\cite{bai2025recammaster} dataset.
Average results of all video sequences are reported.
$\dagger$: To avoid memory overflow, we sample images with interval=8 for VGGT~\cite{wang2025vggt}; by default, other methods use all frames.
The right side shows one reconstruction process of our method.}

\begin{minipage}{0.3\textwidth}
    \centering
    \resizebox{\textwidth}{!}{
        \begin{tabular}{lcccc}
         \toprule
         \textbf{Method} & \textbf{Interval} & {ATE}$\downarrow$ & {RTE}$\downarrow$ & {RRE}$\downarrow$ \\
         \midrule
         COLMAP~\cite{schonberger2016structure} & & 0.073 & \cellcolor{gray2}0.004 & \cellcolor{gray3}0.110 \\
         \hdashline
         \rowcolor{gray!20}
         VGGT~\cite{wang2025vggt} &  & \textit{OOM} & \textit{OOM} & \textit{OOM} \\
         VGGT$^{\dagger}$~\cite{wang2025vggt} & 8 & \cellcolor{gray3}0.027 & 0.016 & 0.177 \\
         Fast3R~\cite{yang2025fast3r} & & 0.144 & 0.056  & 1.948 \\
         FastVGGT~\cite{shen2025fastvggt} & & \cellcolor{gray2}0.023 & \cellcolor{gray3}0.008 & \cellcolor{gray2}0.080 \\
         CUT3R~\cite{wang2025continuous} & & 0.175 & 0.011 & 0.164 \\
         \hdashline
         Ours & & \cellcolor{gray1}{0.005} & \cellcolor{gray1}{0.001} & \cellcolor{gray1}{0.011} \\
         \bottomrule
    \end{tabular}
    }
\end{minipage}
\hfill
\begin{minipage}{0.65\textwidth}
    \centering
    \includegraphics[width=\textwidth]{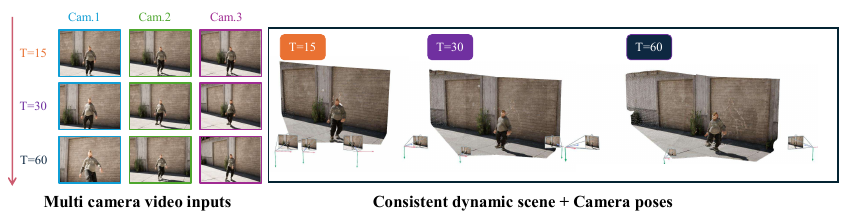}
\end{minipage}

\label{tab:multiplecam_video_cam_pose}
\end{table}
\begin{table}[t]
\centering
\small
\setlength{\tabcolsep}{4pt}

\caption{
Quantitative comparisons of camera trajectories on the \textbf{MultiCamRobolab} dataset.
Average results of all video sequences are reported.
$\dagger$: To avoid memory overflow, we sample images with interval=8 for VGGT~\cite{wang2025vggt};
`$\times$' means fail to reconstruct scenes: not all images are successfully registered.
GPU memory reports the peak inference memory consumption.
}
\label{tab:robotracking}

\resizebox{\linewidth}{!}{%
\begin{tabular}{c|lcccccccccccccc}
\toprule
\multirow{2}{*}{}
& \multirow{2}{*}{\textbf{Method}}
& \multirow{2}{*}{\textbf{Interval}}
& \multicolumn{3}{c}{$\textbf{RoboDog}_{\text{overlap}}$}
& \multicolumn{3}{c}{$\textbf{RoboDog}_{\text{non-overlap}}$}
& \multicolumn{3}{c}{\textbf{RoboArm}}
& \multicolumn{3}{c}{\textbf{DynamicHuman}}
& \multirow{2}{*}{\shortstack{\textbf{GPU Mem.}\\\textbf{(GB)}}} \\
\cmidrule(lr){4-6} \cmidrule(lr){7-9} \cmidrule(lr){10-12} \cmidrule(lr){13-15}
& &
& ATE$\downarrow$ & RTE$\downarrow$ & RRE$\downarrow$
& ATE$\downarrow$ & RTE$\downarrow$ & RRE$\downarrow$
& ATE$\downarrow$ & RTE$\downarrow$ & RRE$\downarrow$
& ATE$\downarrow$ & RTE$\downarrow$ & RRE$\downarrow$
& \\
\midrule
\multicolumn{1}{c|}{} & \multicolumn{15}{l}{\underline{\textbf{Classical SfM}}} \\
a) & COLMAP~\cite{schonberger2016structure} &
& 0.134 & \cellcolor{gray3}0.006 & 0.179
& $\times$ & $\times$ & $\times$
& \cellcolor{gray3}0.008 & \cellcolor{gray2}0.002 & 0.072
& 0.133 & 0.011 & 0.276
& \textbackslash \\
b) & GLOMAP~\cite{pan2024global}
&
& $\times$ & $\times$ & $\times$
& $\times$ & $\times$ & $\times$
& \cellcolor{gray2}0.006 & \cellcolor{gray1}0.001 & \cellcolor{gray1}0.053
& \cellcolor{gray3}0.020 & \cellcolor{gray2}0.009 & 0.259
& \textbackslash \\

\multicolumn{1}{c|}{} & \multicolumn{14}{l}{\cellcolor{gray!20}\textit{with dynamic masks}} \\

c) & COLMAP~\cite{schonberger2016structure} &
& 0.045 & \cellcolor{gray2}0.005 & \cellcolor{gray2}0.174
& $\times$ & $\times$ & $\times$
& \cellcolor{gray3}0.008 & \cellcolor{gray2}0.002 & 0.068
& $\times$ & $\times$ & $\times$
& \textbackslash \\

d) & GLOMAP~\cite{pan2024global} &
& $\times$ & $\times$ & $\times$
& $\times$ & $\times$ & $\times$
& \cellcolor{gray1}0.005 & \cellcolor{gray1}0.001 & \cellcolor{gray2}0.054
& 0.028 & \cellcolor{gray1}0.008 & \cellcolor{gray2}0.250
& \textbackslash \\

\multicolumn{1}{c|}{} & \multicolumn{14}{l}{\cellcolor{gray!20}\textit{with NetVLAD retrieval}} \\

e) & COLMAP~\cite{schonberger2016structure} &
& 0.112 & \cellcolor{gray2}0.005 & \cellcolor{gray3}0.176
& $\times$ & $\times$ & $\times$
& \cellcolor{gray3}0.008 & \cellcolor{gray2}0.002 & 0.071
& 0.082 & \cellcolor{gray3}0.010 & 0.270
& \textbackslash \\

f) & GLOMAP~\cite{pan2024global} &
& $\times$ & $\times$ & $\times$
& $\times$ & $\times$ & $\times$
& \cellcolor{gray1}0.005 & \cellcolor{gray1}0.001 & \cellcolor{gray2}0.054
& \cellcolor{gray2}0.019 & \cellcolor{gray2}0.009 & \cellcolor{gray1}0.244
& \textbackslash \\

\hdashline
\multicolumn{1}{c|}{} & \multicolumn{15}{l}{\underline{\textbf{Feed-Forward}}} \\
g) & VGGT~\cite{wang2025vggt}
&
& \textit{OOM} & \textit{OOM} & \textit{OOM}
& \textit{OOM} & \textit{OOM} & \textit{OOM}
& \textit{OOM} & \textit{OOM} & \textit{OOM}
& \textit{OOM} & \textit{OOM} & \textit{OOM}
& \textit{OOM} \\
h) & VGGT$^{\dagger}$~\cite{wang2025vggt}
& 8
& \cellcolor{gray3}0.024 & 0.012 & 0.446
& \cellcolor{gray1}0.019 & \cellcolor{gray3}0.014 & 0.423
& \cellcolor{gray2}0.006 & 0.004 & 0.187
& 0.032 & 0.057 & 1.713
& 20.45 \\
i) & Fast3R~\cite{yang2025fast3r}
&
& 0.148 & 0.081 & 1.510
& 0.701 & 0.207 & 4.723
& 0.063 & 0.066 & 1.118
& 0.180 & 0.100 & 1.258
& 39.20 \\
j) & FastVGGT~\cite{shen2025fastvggt}
&
& \cellcolor{gray2}0.021 & 0.009 & 0.234
& \cellcolor{gray2}0.020 & \cellcolor{gray2}0.013 & \cellcolor{gray2}0.266
& \cellcolor{gray2}0.006 & 0.004 & 0.107
& \cellcolor{gray3}0.020 & 0.013 & 0.286
& 22.08 \\
k) & CUT3R~\cite{wang2025continuous}
&
& 0.277 & 0.016 & 0.299
& \cellcolor{gray3}0.377 & 0.021 & \cellcolor{gray3}0.326
& 0.055 & \cellcolor{gray3}0.003 & 0.112
& 0.196 & 0.016 & 0.330
& 22.43 \\
\hdashline
l) & Ours
&
& \cellcolor{gray1}0.011 & \cellcolor{gray1}0.003 & \cellcolor{gray1}0.157
& \cellcolor{gray2}0.020 & \cellcolor{gray1}0.003 & \cellcolor{gray1}0.163
& \cellcolor{gray1}0.005 & \cellcolor{gray1}0.001 & \cellcolor{gray3}0.059
& \cellcolor{gray1}0.013 & \cellcolor{gray2}0.009 & \cellcolor{gray3}0.257
& 20.04 \\
\bottomrule
\end{tabular}
}
\end{table}

\begin{table}[t]
\centering
\small
\setlength{\tabcolsep}{4pt}

\caption{
Quantitative comparisons of camera trajectories on the \textbf{MultiCamRobolab-3-cameras dataset}.
}
\label{tab:3-camera-tracking}

\resizebox{0.45\linewidth}{!}{%
    \begin{tabular}{lcccc}
    \toprule
    \textbf{Method}
    & ATE$\downarrow$ & RTE$\downarrow$ & RRE$\downarrow$ \\
    \midrule
    COLMAP~\cite{schonberger2016structure}
    & \cellcolor{gray3}0.343 & \cellcolor{gray2}0.013 & \cellcolor{gray3}0.376\\
    \hdashline
    Fast3R~\cite{yang2025fast3r}
    & 0.376 & 0.154 & 1.825\\
    FastVGGT~\cite{shen2025fastvggt}
    & \cellcolor{gray2}0.032 & \cellcolor{gray3}0.017 & \cellcolor{gray2}0.363 \\
    CUT3R~\cite{wang2025continuous}
    & 0.474 & 0.0351 & 0.467\\
    \hdashline
    Ours
    & \cellcolor{gray1}0.020 & \cellcolor{gray1}0.011 & \cellcolor{gray1}0.326\\
    \bottomrule
    \end{tabular}
}
\vspace{-3pt}
\end{table}

First, we report the \textbf{camera pose evaluation} on the MultiCamVideo dataset and MultiCamRobolab in \cref{tab:multiplecam_video_cam_pose,tab:robotracking,tab:3-camera-tracking}.
Our method achieves the best results in the MultiCamVideo datasets, as shown in \cref{tab:multiplecam_video_cam_pose}.
The right side in \cref{tab:multiplecam_video_cam_pose} shows the visualization of one reconstruction process of our method.
Our method achieves the overall better results in MultiCamRobolab datasets (shown in \cref{tab:robotracking,tab:3-camera-tracking}).
Specifically, the classical SfM method COLMAP~\cite{schonberger2016structure} tends to produce noisy pose estimates in the presence of dynamic objects.
While GLOMAP~\cite{pan2024global} yields more accurate estimates, it fails to reconstruct two of the four scenes.
Interestingly, we find that pruning dynamic objects does not always improve reconstruction.
For example, on the DynamicHuman scene, comparing (a),(b) with (c),(d) shows that applying dynamic masks instead degrades performance.
We attribute this to three factors:
(i) dynamic objects are momentarily \emph{statically consistent} across synchronized cameras, providing valid inter-camera correspondences;
(ii) at 30~FPS, object motion is small, so features can remain locally consistent over short time windows; and
(iii) in textureless environments, dynamic objects supply critical keypoints.
In summary, dynamic objects can themselves contribute to image matching, and removing them simply reduces the number of available correspondences.
Unlike these baselines, our method relies on dense neural optical flow; through the weighting mechanism in our optimization, it implicitly accounts for correspondence reliability by down-weighting outliers that violate multi-view consistency.
COLMAP and GLOMAP can also fail in the non-overlap scenes (see the $\rm{RoboDog}_{\text{non-overlap}}$ in \cref{tab:robotracking}).
VGGT cannot process all frames, even on a single A100-40GiB machine.
For other feed-forward models, Fast3R also can not handle dynamic objects well; though CUT3R is designed for processing videos, it can not achieve satisfactory results (see \cref{fig_traj_vis}, for each separate video clip, CUT3R can not generate smooth results).
In comparison, FastVGGT's results are the best in all feed-forward models, which reveals the advantage of processing all frames together instead of test-time optimization used in CUT3R.
In addition, the results indicate that FastVGGT is robust to dynamic objects in the scene. This observation motivates further investigation into the conditions under which such reconstruction models succeed or fail in dynamic environments.
Our method achieves the best overall performance across all scenes (while consuming the least GPU memory), except for $\rm{RoboDog}_{\text{non-overlap}}$. In this case, the cameras have few overlapping fields of view, causing the spatio-temporal connections to degenerate into purely temporal connections within each individual camera stream (more discussions can be found in the ablation study~\cref{subsec:ablation_study}).
\begin{figure}[t]
    \centering

    \begin{subfigure}[b]{0.56\linewidth}
        \centering
        \begin{minipage}{0.48\linewidth}
            \centering
            \includegraphics[width=\linewidth]{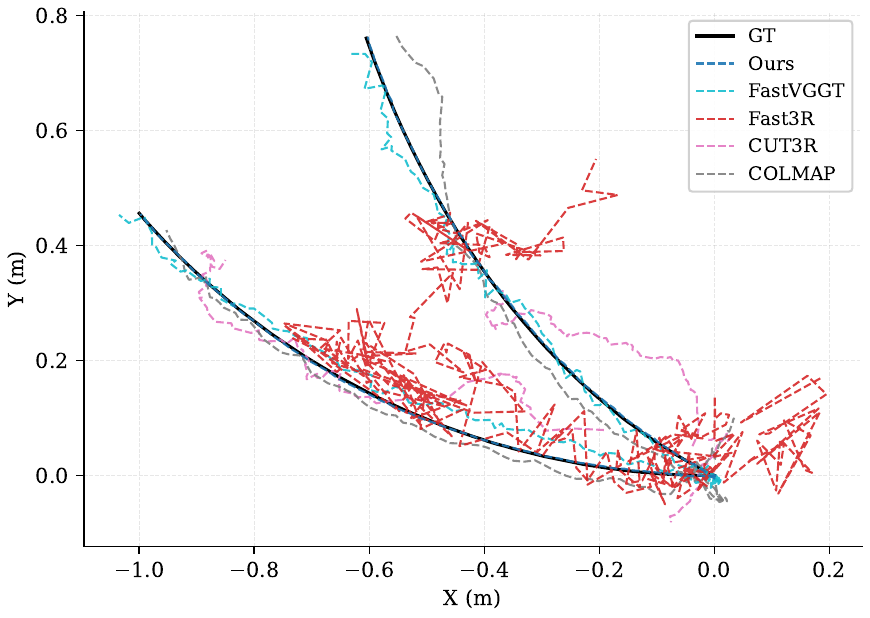}
        \end{minipage}
        \hfill
        \begin{minipage}{0.48\linewidth}
            \centering
            \includegraphics[width=\linewidth]{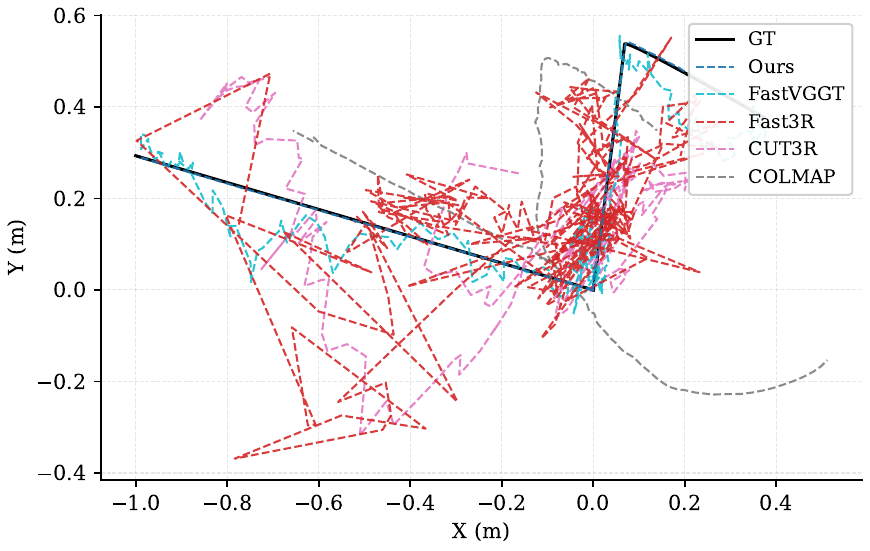}
        \end{minipage}
        \caption{MultiCamVideo}
        \label{fig:sub_mvv}
    \end{subfigure}
    \hfill
    \begin{subfigure}[b]{0.40\linewidth}
        \centering
        \begin{minipage}{0.48\linewidth}
            \centering
            \includegraphics[width=\linewidth]{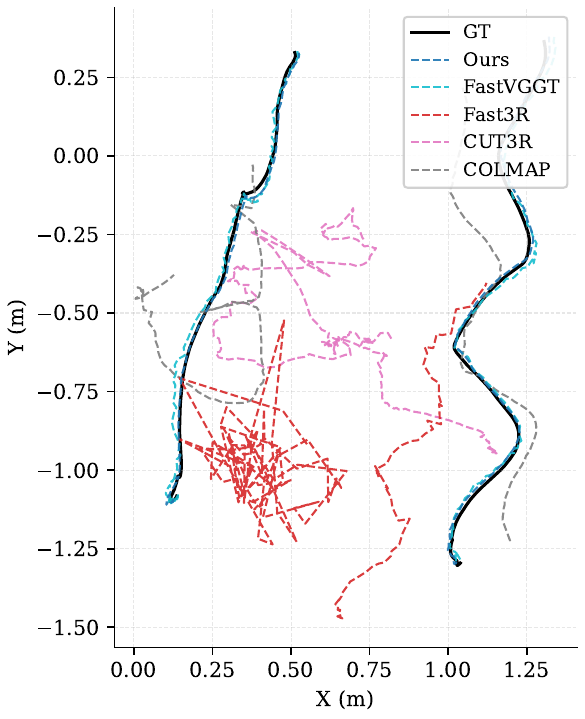}
        \end{minipage}
        \hfill
        \begin{minipage}{0.48\linewidth}
            \centering
            \includegraphics[width=\linewidth]{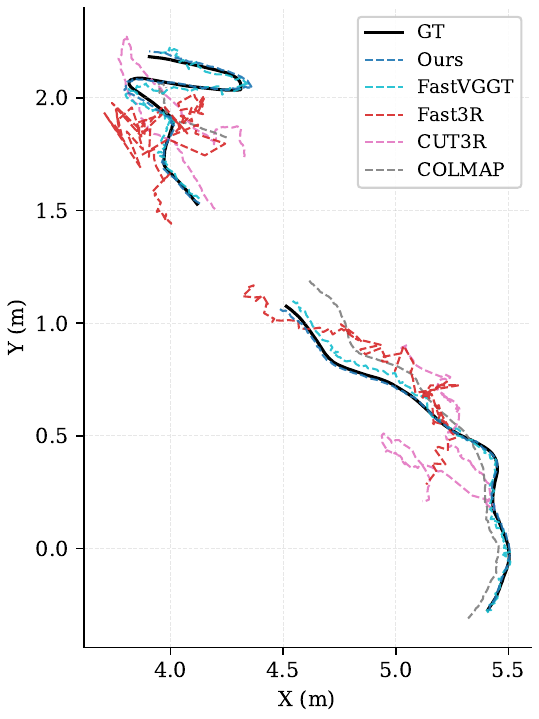}
        \end{minipage}
        \caption{MultiCamRobolab}
        \label{fig:sub_robolab}
    \end{subfigure}

    \caption{
    Visualization (projected to X-Y plane) of camera trajectories estimated by different methods in the two datasets.
    Multiple camera trajectories are treated as one and aligned with GT trajectories by SIM(3) alignment.}
    \label{fig_traj_vis}
    \vspace{-4pt}
\end{figure}
The visualization of estimated trajectories compared to ground-truth can be found in \cref{fig_traj_vis}.

Second, the \textbf{depth and scene consistency evaluation} is shown in \cref{tab:scene_comparison}. Since the MultiCamVideo does not provide ground-truth depth, we report quantitative evaluation results only on the MultiCamRobolab dataset in \cref{tab:scene_comparison}.
Our method combines the prior depth and spatio-temporal connection refinement, achieving the most consistent depth evaluation.
Notably, the scene consistency evaluation in fact combines both pose evaluation and depth evaluation.
Though methods such as CUT3R achieve good results on single-frame video depth evaluation, they can not generate consistent scene results.
The qualitative reconstruction results can be seen in \cref{fig:robolab_scene_vis}.

\begin{table}[t]
\centering
\small
\setlength{\tabcolsep}{3pt}
\caption{
Quantitative comparisons of frame depth and scene consistency on the \textbf{MultiCamRobolab} dataset.
    Average results of all video sequences are reported.
    $\dagger$:To avoid memory overflow, we sample images with interval=8 for VGGT~\cite{wang2025vggt}; by default, other methods use all frames.}
\label{tab:scene_comparison}
\resizebox{\textwidth}{!}{%
\begin{tabular}{lccccccccccccc}
\toprule
\multirow{2}{*}{\textbf{Method}} & \multirow{2}{*}{\textbf{Interval}} &
\multicolumn{3}{c}{$\textbf{RoboDog}_{\text{overlap}}$} &
\multicolumn{3}{c}{$\textbf{RoboDog}_{\text{non-overlap}}$} &
\multicolumn{3}{c}{\textbf{RoboArm}} & \multicolumn{3}{c}{\textbf{DynamicHuman}} \\
\cmidrule(lr){3-5} \cmidrule(lr){6-8} \cmidrule(lr){9-11} \cmidrule(lr){12-14}
& & Abs.Rel$\downarrow$ & $\delta_{1.25}$ $\uparrow$ & $M_d$$\downarrow$
& Abs.Rel$\downarrow$ & $\delta_{1.25}$ $\uparrow$ & $M_d$$\downarrow$
& Abs.Rel$\downarrow$ & $\delta_{1.25}$ $\uparrow$ & $M_d$$\downarrow$
& Abs.Rel$\downarrow$ & $\delta_{1.25}$ $\uparrow$ & $M_d$$\downarrow$\\
\midrule
VGGT~\cite{wang2025vggt} &  & \textit{OOM} & \textit{OOM} & \textit{OOM} & \textit{OOM} &
\textit{OOM} & \textit{OOM} & \textit{OOM} & \textit{OOM} & \textit{OOM} & \textit{OOM} & \textit{OOM} & \textit{OOM} \\
VGGT$^{\dagger}$~\cite{wang2025vggt} & 8 & \cellcolor{gray3}0.068 & 0.939 & \cellcolor{gray3}0.376 & \cellcolor{gray3}0.060 & 0.947 & \cellcolor{gray3}0.327 & 0.171 & \cellcolor{gray3}0.772 & \cellcolor{gray3}0.547 & \cellcolor{gray2}0.194 & \cellcolor{gray3}0.900 & \cellcolor{gray3}0.461\\
Fast3R~\cite{yang2025fast3r} &  & 0.144 & 0.788 & 0.580 & 0.157 & 0.749 & 1.501 &0.191 & 0.714 & 0.763 & 0.214 & 0.735 & 1.799\\
FastVGGT~\cite{shen2025fastvggt} & & \cellcolor{gray2}0.026 & \cellcolor{gray2}0.978 & \cellcolor{gray2}0.128 & \cellcolor{gray1}0.018 & \cellcolor{gray2}0.988 & \cellcolor{gray2}0.095 & \cellcolor{gray2}0.060 & \cellcolor{gray2}0.906 & \cellcolor{gray2}0.292 & \cellcolor{gray1}0.030 & \cellcolor{gray1}0.987 & \cellcolor{gray2}0.185\\
CUT3R~\cite{wang2025continuous} & & 0.048 & \cellcolor{gray3}0.968 & 0.715 & \cellcolor{gray3}0.040 & \cellcolor{gray3}0.987 & 0.581 &\cellcolor{gray3}0.158 & 0.745 & 0.708 & \cellcolor{gray3}0.060 & \cellcolor{gray3}0.963 & 0.503\\
\hdashline
Ours & & \cellcolor{gray1}0.011 & \cellcolor{gray1}0.989 & \cellcolor{gray1}0.091 &
\cellcolor{gray1}0.018 & \cellcolor{gray1}0.991 & \cellcolor{gray1}0.092 &
\cellcolor{gray1}0.059 & \cellcolor{gray1}0.947 & \cellcolor{gray1}0.289 & \cellcolor{gray1}0.030 & \cellcolor{gray2}0.971 & \cellcolor{gray1}0.112\\
\bottomrule
\end{tabular}
}
\end{table}

\begin{figure}[th]
    \centering

    \newcommand{\vinw}{0.12\textwidth}   
    \newcommand{\mew}{0.25\textwidth}   
    \newcommand{\rowsep}{0.35pt}         

    \newcommand{\imgcell}[2]{%
        \begin{minipage}[c]{#1}\centering
            \includegraphics[width=\linewidth]{#2}
        \end{minipage}
    }

    \newcommand{\videoplaceholder}[1]{%
        \begin{minipage}[c]{\vinw}\centering
            \fbox{\parbox[c][#1][c]{0.95\linewidth}{\centering \scriptsize Input video}}
        \end{minipage}
    }

    \setlength{\tabcolsep}{3pt}  
    \renewcommand{\arraystretch}{1.0}
    \resizebox{\textwidth}{!}{%
      \begin{tabular}{@{}c|cccc@{}}
        \multicolumn{1}{c}{\begin{minipage}[b]{\vinw}\centering \scriptsize \mbox{Input videos}\end{minipage}} &
        \multicolumn{1}{c}{\begin{minipage}[b]{\mew}\centering \scriptsize Ours\end{minipage}} &
        \multicolumn{1}{c}{\begin{minipage}[b]{\mew}\centering \scriptsize FastVGGT~\cite{shen2025fastvggt}\end{minipage}} &
        \multicolumn{1}{c}{\begin{minipage}[b]{\mew}\centering \scriptsize CUT3R~\cite{wang2025continuous}\end{minipage}} &
        \multicolumn{1}{c}{\begin{minipage}[b]{\mew}\centering \scriptsize Fast3R~\cite{yang2025fast3r}\end{minipage}} \\

        \imgcell{\vinw}{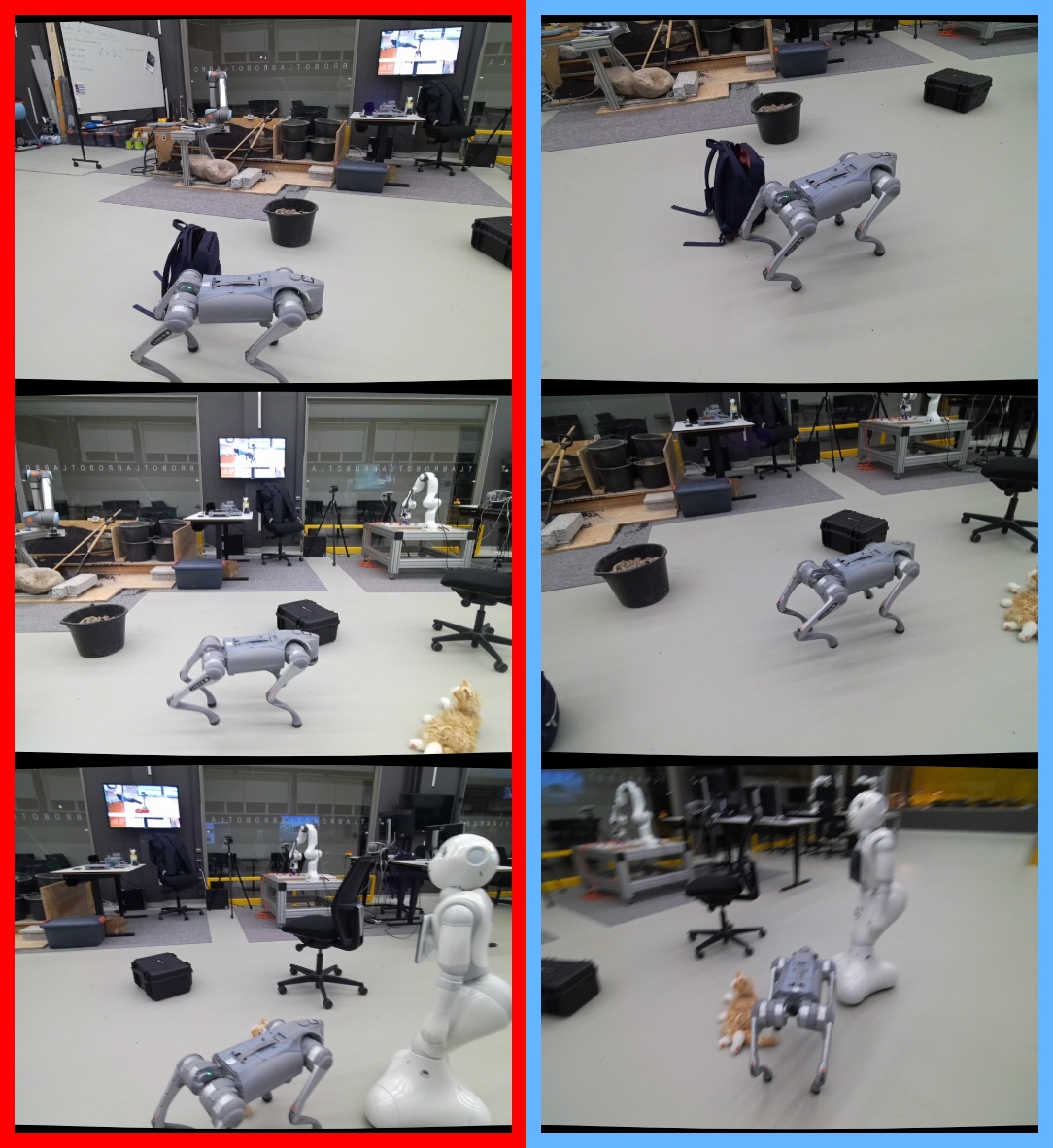} &
        \imgcell{\mew}{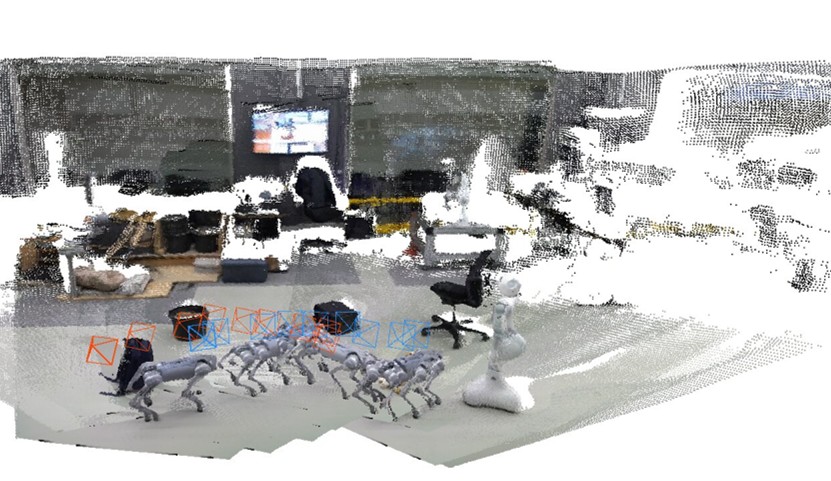} &
        \imgcell{\mew}{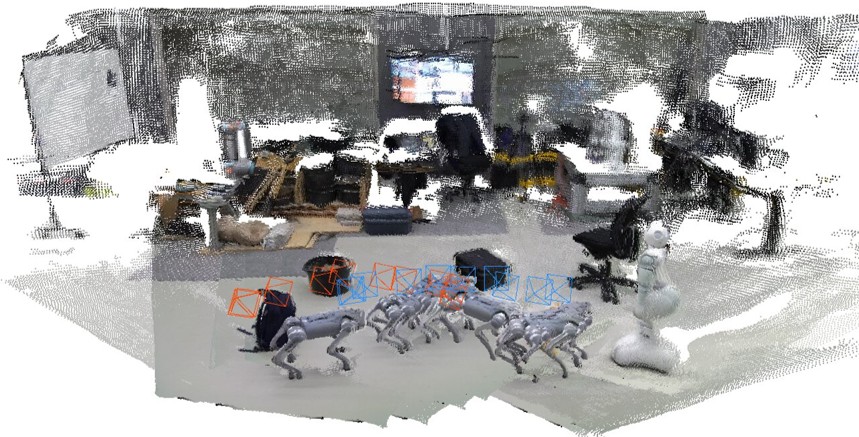} &
        \imgcell{\mew}{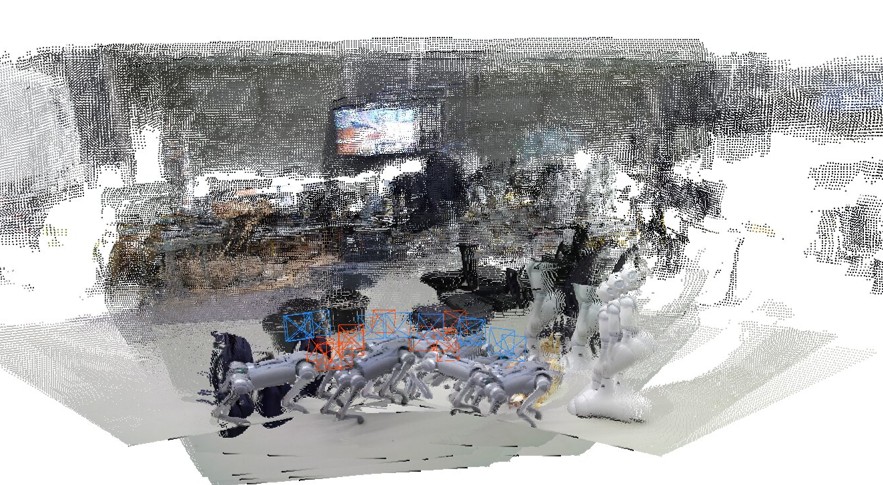} &
        \imgcell{\mew}{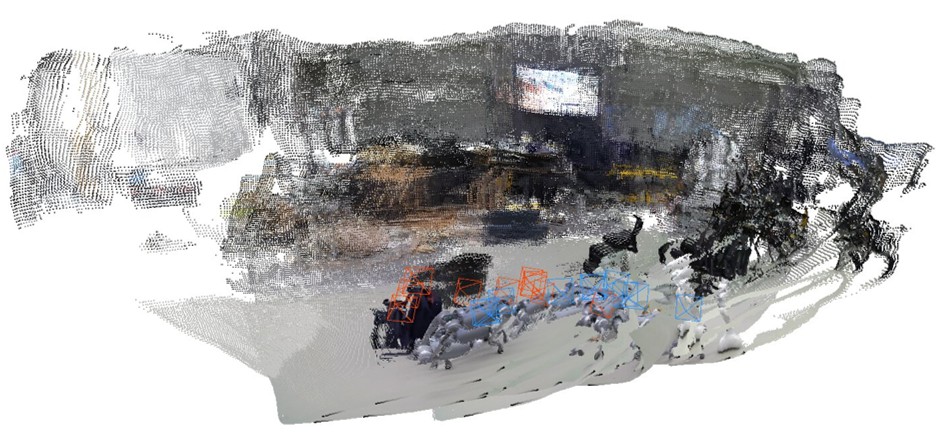} \\
        \addlinespace[2pt]

        {\scriptsize \ldots} \\

        \imgcell{\vinw}{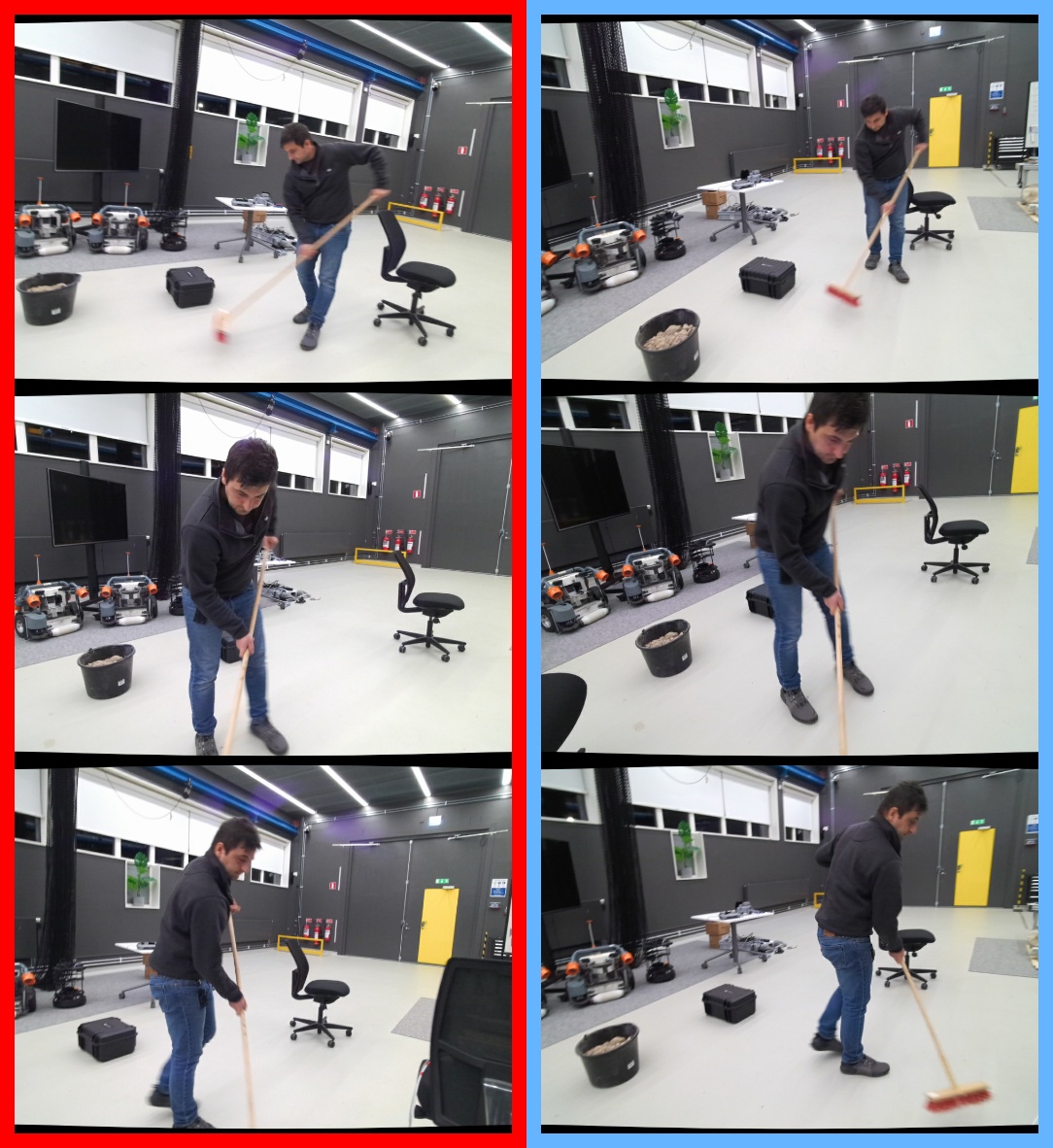} &
        \imgcell{\mew}{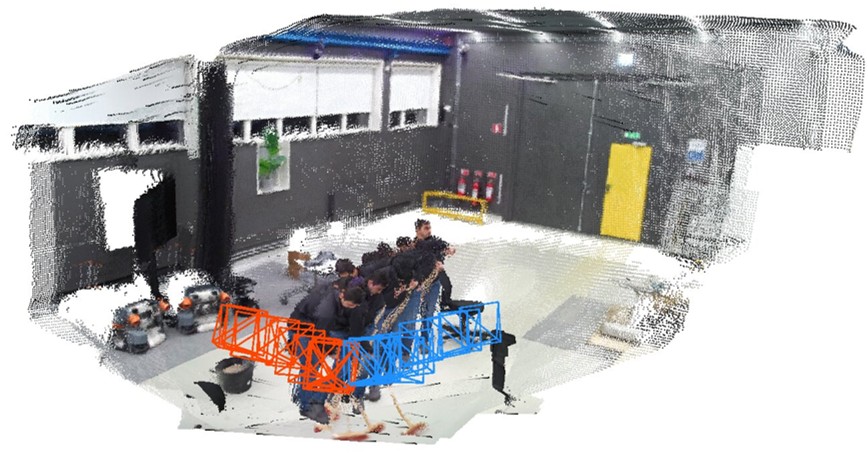} &
        \imgcell{\mew}{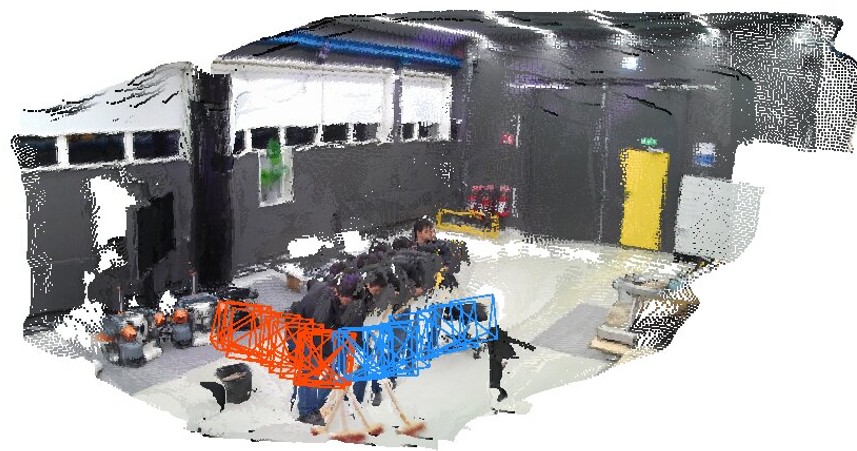} &
        \imgcell{\mew}{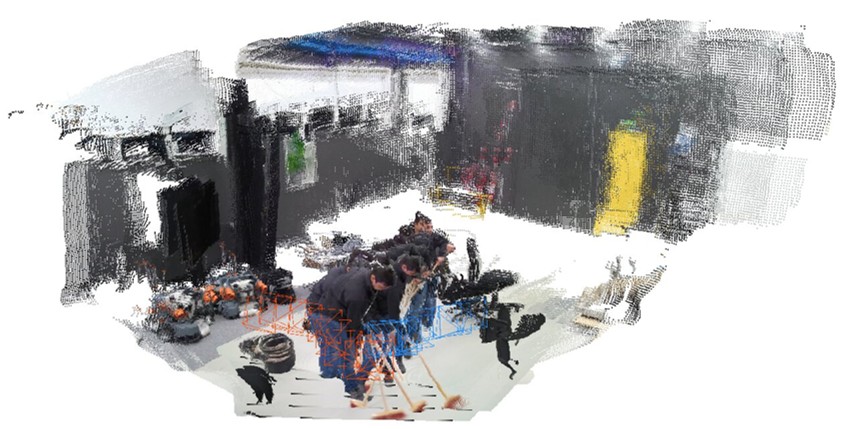} &
        \imgcell{\mew}{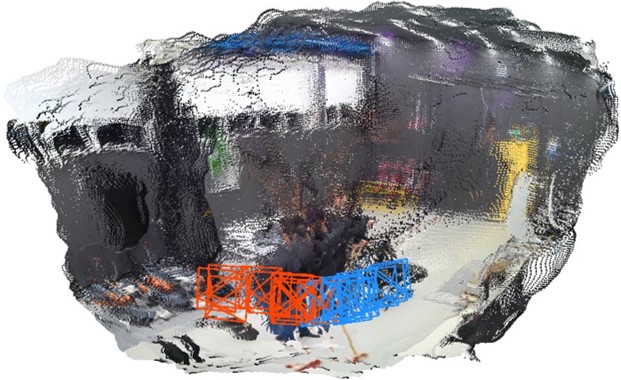} \\
      \end{tabular}%
    }

    \caption{\textbf{Qualitative reconstruction results on MultiCamRobolab datasets.}}
    \label{fig:robolab_scene_vis}
    \vspace{-4pt}
\end{figure}

\subsection{Ablation Studies}
\label{subsec:ablation_study}

\begin{table}[t]
\centering

\caption{
\textbf{Ablation studies on camera pose tracking.}
The \enquote{\texttt{Full method}} is used as the baseline, and all other results are reported as relative changes with respect to this baseline.
{\color{red}Red} entries indicate performance degradation, whereas {\color{blue}blue} entries indicate performance improvement.
}

\resizebox{\linewidth}{!}{%
\begin{tabular}{lcccccccccccc}
\toprule
\multirow{2}{*}{\textbf{Method}}
    & \multicolumn{3}{c}{$\textbf{RoboDog}_{\text{overlap}}$}
    & \multicolumn{3}{c}{$\textbf{RoboDog}_{\text{non-overlap}}$}
    & \multicolumn{3}{c}{\textbf{RoboArm}}
    & \multicolumn{3}{c}{\textbf{DynamicHuman}} \\
\cmidrule(lr){2-4} \cmidrule(lr){5-7} \cmidrule(lr){8-10} \cmidrule(lr){11-13}
    & ATE$\downarrow$ & RTE$\downarrow$ & RRE$\downarrow$
    & ATE$\downarrow$ & RTE$\downarrow$ & RRE$\downarrow$
    & ATE$\downarrow$ & RTE$\downarrow$ & RRE$\downarrow$
    & ATE$\downarrow$ & RTE$\downarrow$ & RRE$\downarrow$ \\
\midrule
$\texttt{w/o}$ W.B. Init.
    & \textcolor{red!90}{+0.284} & \textcolor{red!50}{+0.003} & \textcolor{blue!40}{-0.002}
    & \textcolor{red!100}{+0.496} & \textcolor{red!70}{+0.019} & \textcolor{red!55}{+0.026}
    & \textcolor{red!80}{+0.084} & \textcolor{red!45}{+0.002} & \textcolor{red!40}{+0.001}
    & \textcolor{red!85}{+0.130} & \textcolor{red!40}{+0.001} & \textcolor{red!45}{+0.009}\\
$\texttt{w/o}$ ST-Graph
    & \textcolor{red!75}{+0.158} & \textcolor{red!50}{+0.003} & \textcolor{blue!40}{-0.002}
    & \textcolor{red!20}{+0.006} & \textcolor{black!40}{+0.000} & \textcolor{blue!40}{-0.002}
    & \textcolor{red!50}{+0.012} & \textcolor{black!40}{+0.000} & \textcolor{blue!35}{-0.001}
    & \textcolor{red!85}{+0.141} & \textcolor{red!45}{+0.002} & \textcolor{red!45}{+0.007} \\
$\texttt{Full method}$
    & 0.011 & 0.003 & 0.157
    & 0.020 & 0.003 & 0.163
    & 0.005 & 0.001 & 0.059
    & 0.013 & 0.009 & 0.257 \\
\bottomrule
\end{tabular}
}
\label{tab:abliation_study_pose_tracking}

\vspace{6pt}
\setlength{\tabcolsep}{2pt}
\caption{\textbf{Ablation studies on noisy initializaztion.}}
\resizebox{0.20\linewidth}{!}{%
\begin{tabular}{l|c}
\toprule
\textbf{Noise level} & \textbf{ATE}$\downarrow$\\
\midrule
0 & 0.013 \\
0.02 m& 0.013 \\
0.05 m& 0.016 \\
0.10 m& 0.044 \\
\bottomrule
\end{tabular}
}
\label{tab:abliation_study_right}
\end{table}
In this section, we perform ablation studies to evaluate the contributions of the key components of our method. In particular, we investigate the effectiveness of (1) the proposed wide-baseline initialization, (2) the spatio-temporal graph used for camera pose tracking, and (3) the refinement stage for improving depth estimation and scene consistency.
In \cref{tab:abliation_study_pose_tracking}, we report the trajectory differences obtained by removing the wide-baseline initialization~(\hyperref[para:init]{W.B. Init.}), and the spatio-temporal connection~(\hyperref[para:spatio-temporal]{ST-Graph}). The full method is treated as the baseline, and all other results are presented as deviations relative to the full method. {\color{red}Red} indicates performance degradation, whereas {\color{blue}blue} denotes performance improvement.
By the results, we can see that the proposed wide-baseline initialization is important for the tracking.
That is because, unlike traditional single-camera tracking problems, there is often not enough overlap across different cameras for initialization; it is important to use some reconstruction model to get a prior estimation.
But our method is also robust to noisy initialization; we conduct an extra ablation study by adding noise to the initialized poses.
In \cref{tab:abliation_study_right}, we add rotational noise of $3^{\circ}$ and different levels of translation noise; the results show that our method can recover from the noisy initialized poses under mild conditions.
Note that in \cref{tab:abliation_study_pose_tracking}, ``$\texttt{w/o}$ {ST-Graph}'' means our method degenerates to running separate video reconstruction and then aligning them via feed-forward predicted poses.
The second row in \cref{tab:abliation_study_pose_tracking} shows that the spatio-temporal graph can help improve the tracking results because it introduces more constraints during the tracking.
Note that for the scene $\rm{RoboDog}_{\text{non-overlap}}$, the spatio-temporal graph helps least, because there is little overlap across cameras.
In \cref{tab:abliation_study_depth}, we show that optimizing the per-depth-frame scale and offset is not enough; two phases of refinement can help improve the depth accuracy and scene consistency.
(More ablation studies on refinement and spatio-temporal connection edge counts can be found in the supplementary materials.)
\begin{table}[ht]
\centering
\small
\setlength{\tabcolsep}{4pt}
\renewcommand{\arraystretch}{1.1}
\caption[]{
\textbf{Ablation studies on video depth evaluation and scene consistency.}
We evaluate the effect of the refinement proposed in \cref{subsec:refinement}.
The results show that the refinement stage can improve overall scene consistency.
}
\resizebox{\linewidth}{!}{%
\begin{tabular}{ccccccccccccc}
\toprule
\multirow{2}{*}{\textbf{Method}}
    & \multicolumn{3}{c}{$\textbf{RoboDog}_{\text{overlap}}$}
    & \multicolumn{3}{c}{$\textbf{RoboDog}_{\text{non-overlap}}$}
    & \multicolumn{3}{c}{\textbf{RoboArm}} & \multicolumn{3}{c}{\textbf{DynamicHuman}} \\
\cmidrule(lr){2-4} \cmidrule(lr){5-7} \cmidrule(lr){8-10} \cmidrule(lr){11-13}
    & {Abs.Rel}$\downarrow$ & {$\delta_{1.25}$} $\uparrow$ & {$M_d$}$\downarrow$
    & {Abs.Rel}$\downarrow$ & {$\delta_{1.25}$} $\uparrow$ & {$M_d$}$\downarrow$
    & {Abs.Rel}$\downarrow$ & {$\delta_{1.25}$} $\uparrow$ & {$M_d$}$\downarrow$
    & {Abs.Rel}$\downarrow$ & {$\delta_{1.25}$} $\uparrow$ & {$M_d$}$\downarrow$\\
\midrule
\textcolor{red}{$\times$} {Phase1}; \textcolor{red}{$\times$} {Phase2}& \textcolor{red!80}{+0.020} & \textcolor{blue!20}{+0.001} & \textcolor{red!90}{+0.066} & \textcolor{red!70}{+0.014} & \textcolor{blue!20}{+0.001} & \textcolor{red!85}{+0.060} & \textcolor{red!90}{+0.068} & \textcolor{red!100}{-0.123} & \textcolor{red!100}{+0.226} & \textcolor{red!80}{+0.028} & \textcolor{blue!20}{+0.007} & \textcolor{red!80}{+0.063}\\
\textcolor{green}{\checkmark} {Phase1}; \textcolor{red}{$\times$} {Phase2}& \textcolor{red!70}{+0.012} & \textcolor{blue!20}{+0.001} & \textcolor{red!60}{+0.029} & \textcolor{red!50}{+0.005} & \textcolor{blue!20}{+0.001} & \textcolor{red!50}{+0.012} & \textcolor{red!75}{+0.044} & \textcolor{red!85}{-0.083} & \textcolor{red!75}{+0.087} & \textcolor{red!20}{+0.003} & \textcolor{blue!20}{+0.007} & \textcolor{red!40}{+0.012}\\
\textcolor{green}{\checkmark} {Phase1}; \textcolor{green}{\checkmark} {Phase2} & 0.011 & 0.989 & 0.091 & 0.018 & 0.991 &  0.092 & 0.059 & 0.947 & 0.289 & 0.030  & 0.971 & 0.112\\
\bottomrule
\end{tabular}
}
\label{tab:abliation_study_depth}
\vspace{-4pt}
\end{table}

\section{Conclusion}
\label{sec:conclusion}
We introduce the first framework for dense dynamic scene reconstruction and camera pose estimation from multiple free-moving cameras.
We use the feed-forward reconstruction model for robust initialization and introduce a spatio-temporal connection graph for multi-camera tracking in a consistent way.
Based on the constructed connection graph, we introduce a framework for optimizing depths across multiple cameras.
Our method achieves better results while consuming less GPU memory compared to the state-of-the-art feed-forward reconstruction models.

\fi

\bibliographystyle{splncs04}
\bibliography{main}
\end{document}